\pdfoutput=1
\documentclass[letterpaper, 10 pt, conference]{ieeeconf}  

\IEEEoverridecommandlockouts                              

\usepackage{subfiles}
\usepackage{amsmath}
\usepackage{amssymb}
\usepackage{comment}
\usepackage{siunitx}
\usepackage{booktabs} 
\usepackage[table,dvipsnames]{xcolor}
\usepackage{etoolbox}
\usepackage[hidelinks]{hyperref}

\makeatletter
\patchcmd{\@makecaption}
  {\scshape}
  {}
  {}
  {}
\makeatother

\definecolor{rowblue}{RGB}{220,230,240}

\usepackage{chngcntr}
\counterwithin{paragraph}{subsection} 
\usepackage[footnotesize]{caption}
\usepackage{pgfplots} 
\usepackage{pgfplotstable} 
\pgfplotsset{compat=newest} 
\usetikzlibrary{plotmarks} 
\usetikzlibrary{colorbrewer} 
\definecolor{olivegreen}{RGB}{0,170,0}
\definecolor{tealblue}{RGB}{20,100,200}
\newcommand{\argmin}{\operatornamewithlimits{argmin}}
\newcommand{\norm}[1]{\left\lVert#1\right\rVert}
\title{\LARGE \bf
Automatic Tool Landmark Detection for Stereo Vision in Robot-Assisted Retinal Surgery
}

\author{Thomas Probst$^{*1}$, Kevis-Kokitsi Maninis$^{*1}$, Ajad Chhatkuli$^{1}$, Mouloud Ourak$^{2}$, Emmanuel Vander Poorten$^{2}$, and \\ 
Luc Van Gool$^{1,2}$
\thanks{$^{*}$ First two authors have equal contribution.}
\thanks{$^{1}$Thomas Probst, Kevis-Kokitsi Maninis, and Ajad Chhatkuli are with Computer Vision Lab, ETH Z\"urich, Switzerland.\newline
        {\tt\scriptsize \{probstt, kmaninis, ajad.chhatkuli\}@vision.ee.ethz.ch}}%
\thanks{$^{2}$Mouloud Ourak, Emmanuel Vander Poorten are with KU Leuven, Belgium.}
\thanks{$^{1,2}$Luc Van Gool is with Computer Vision Lab, ETH Z\"urich, Switzerland and KU Leuven, Belgium.}
\thanks{The data used in this paper are released at {\scriptsize \url{http://www.vision.ee.ethz.ch/\~kmaninis/keypoints2stereo/}}%
}}

\begin{document}

\maketitle
\thispagestyle{empty}
\pagestyle{empty}

\begin{abstract}

Computer vision and robotics are being increasingly applied in medical interventions. Especially in interventions where extreme precision is required they could make a difference. One such application is robot-assisted retinal microsurgery. In recent works, such interventions are conducted under a stereo-microscope, and with a robot-controlled surgical tool. The complementarity of computer vision and robotics has however not yet been fully exploited. In order to improve the robot control we are interested in 3D reconstruction of the anatomy and in automatic tool localization using a stereo microscope. In this paper, we solve this problem for the first time using a single pipeline, starting from uncalibrated cameras to reach metric 3D reconstruction and registration, in retinal microsurgery. The key ingredients of our method are: (a) surgical tool landmark detection, and (b) 3D reconstruction with the stereo microscope, using the detected landmarks.
To address the former, we propose a novel deep learning method that detects and recognizes keypoints in high definition images at higher than real-time speed. We use the detected 2D keypoints along with their corresponding 3D coordinates obtained from the robot sensors to calibrate the stereo microscope using an affine projection model. We design an online 3D reconstruction pipeline that makes use of smoothness constraints and performs robot-to-camera registration. The entire pipeline is extensively validated on open-sky porcine eye sequences. Quantitative and qualitative results are presented for all steps.
\end{abstract}


\section{Introduction}

Robot and computer vision-assisted surgical procedures are becoming more and more popular due to their ability to attain high precision.
One such procedure in opthalmology involves the peeling of a retinal membrane to improve human vision. In this setup, the surgeon observes the retina and the tool under a stereo microscope while using a robotic arm to control the surgical tool with high precision. This work also builds on such setup, consisting of a surgical tool which is positioned by a robot, and a stereo camera pair that is directly mounted on the surgical microscope. Generally, in such a setup the position of the surgical tool is known with respect to the robot's reference frame, but its position relative to the retinal surface and the cameras is unknown. As a result, for the robot to safely operate in an allowed region inside the eye, additional distance sensors are used to measure and maintain a safe distance to the retina. Visual guidance, however, still remains infeasible due to the different camera and robot coordinate systems. This means that information that comes from the processed images, e.g.,\ the outcome of a vessel segmentation algorithm~\cite{Man+16}, cannot be effectively used. Due to limitations of the microscope acquisition it is further difficult to recover the actual 3D retinal surface. Therefore, accurate localization of the tool with respect to the retinal surface at every instant during surgery remains a very challenging problem.

\begin{figure*}[ht]
\centering
\includegraphics[width=0.85\textwidth]{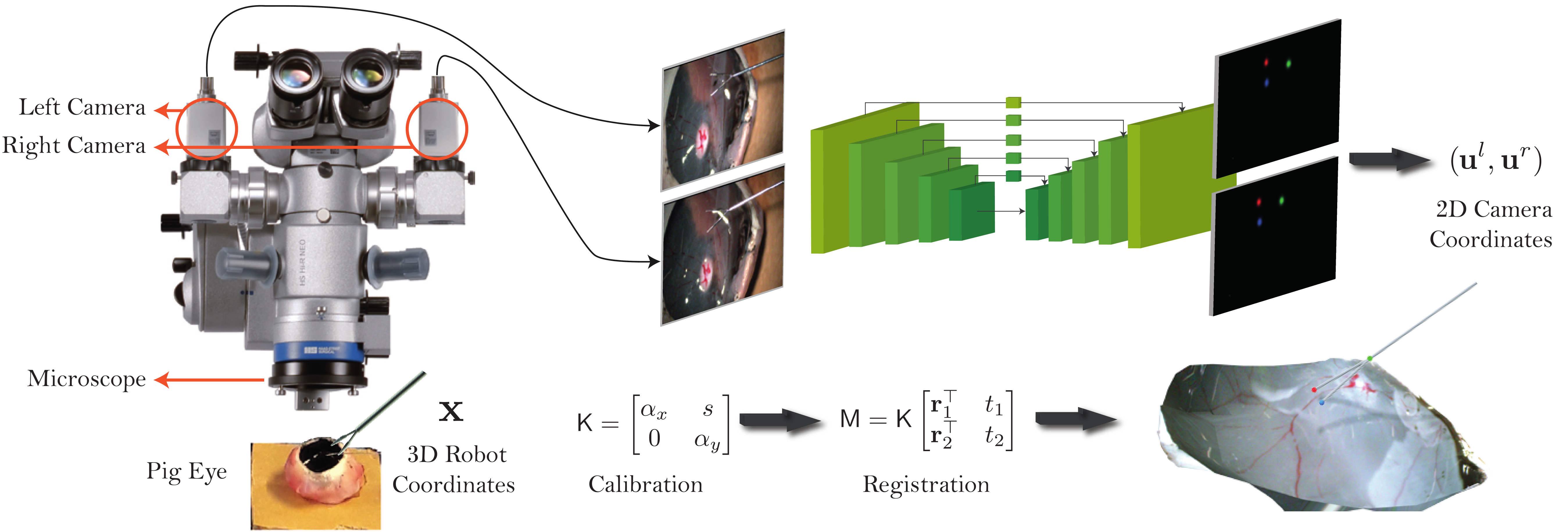}
\vspace{-1mm}
\caption{\textbf{Overview of our method}. Proposed pipeline for stereo calibration, retina reconstruction and tool registration. The microscope and the object of interest are not shown at the correct, relative scale. In the actual setup, the object is orders of times smaller and farther from the microscope.}
\label{fig:pipeline}
\vspace{-6mm}
\end{figure*}

In this paper we tackle the problem of stereo microscope calibration, 3D reconstruction of the retina, and the registration of the landmark points on the tool with respect to the retinal surface. This is the first time all these problems are tackled together. In order to localize both the tool and the retinal surface in 3D, we exploit the robot kinematics which can be measured very accurately with current robotic systems. In this context we solve two important vision problems online: detecting the tool points accurately in the images and reconstructing the retina and the tool points in scale using the stereo microscope camera. Both are challenging problems on their own~\cite{Sznitman2013,Rieke2015,Chanwimaluang2006}. Detecting tool landmark points requires to take into account changes in viewpoint, defocused images, specularities and fast movements of the tool. In addition, the surface reconstruction problem is hindered by the difficulty of calibrating the microscope cameras and the specularities in the images. Unlike consumer cameras, microscope cameras used in retinal microsurgery pose additional challenges: \emph{a)} the narrow field of view and very long effective focal length \emph{b)} a small depth of field, \emph{c)} rolling shutter and \emph{d)} varying rotation and unknown baseline. These challenges make calibration very difficult in practice. Additionally, it is not obvious which camera model and reconstruction strategy best fits the problem of stereo reconstruction from microscopes used in retinal microsurgery. We show that automatically detecting tool landmarks in images, together with their respective 3D positions as they are directly obtained from the robot kinematics provides a reliable solution for microscope calibration and for the retinal reconstruction and tool registration. Figure~\ref{fig:pipeline} gives an overview of our method.

The first task we tackle is markerless surgical tool keypoint detection in images. Several methods~\cite{Bouget2017} have been developed to detect tools in images for various types of surgery. Most are based on hand-crafted features such as those obtained by color image transforms, image gradients and/or RGB pixel intensities \cite{Lee1994,Allan2013,Chhatkuli2014}.
Some work looked into tool detection in the specific case of retinal microsurgery~\cite{Sznitman2013,Rieke2015}. Such previous contributions were either restricted to generation of rather inaccurate bounding boxes, or their computational cost precluded real-time applications. We cast surgical tool detection as a landmark localization problem. We draw inspiration from deep learning algorithms initially used for human pose estimation, in order to detect the tool landmark points in images. We obtain automatically the 3D to 2D correspondences of these landmark keypoints from the robot kinematics. Our detection method runs on full-HD resolution ($1920 \times 1080$ pixels) without the use of markers, at a frame rate of 35 frames per second with a GPU, and requires very few examples of annotated images for training. Experiments show the effect of image resolution on the detection of the tool landmarks and how detection noise affects the camera calibration.

We tackle the problem of stereo microscope calibration using the detected tool landmark points, by assuming a full affine camera model~\cite{Hartley2003} for each microscope camera. Previous methods~\cite{Chanwimaluang2006,Horaud1997} propose calibration of the affine camera by first reconstructing the object with affine Structure from Motion (SfM)~\cite{Tomasi1992} and then computing the suitable upgrade for calibration. This may not be reliable due to the inherent problems of affine factorization-based SfM with respect to noise, the occurrence of missing data, and reconstruction ambiguities. In contrast, we formulate the calibration independent of the reconstruction, and base it solely on the measured robot motion. This frees the calibration from potential errors in factorization based reconstruction.
In our pipeline, the tool is first moved around under the fixed cameras such that a few depths are covered, while the 3D positions from the robot encoder and the observed landmark 2D detections are recorded. We then use the full affine camera model to calibrate the intrinsics as well as the extrinsics using a Gaussian noise prior on the measurements and affine bundle adjustment. In order to initialize the bundle adjustment we use the Direct Linear Transform (DLT)~\cite{Hartley2003}. The projection matrices obtained from the DLT calibration can be directly used to triangulate any stereo correspondence to a 3D point in the robot reference coordinates at the correct scale. We reconstruct the retinal surface by fitting a single smooth surface to the triangulated points. We use Bicubic B-Splines (BBS) to estimate the surface, using the point cloud while catering for its outliers and noise. To the best of our knowledge, this is the first work to employ the calibrated affine camera model for triangulating stereo pair image correspondences with tens of $\mu m$ accuracy. This is an important result as calibration based on checkerboard patterns~\cite{Zhang2000} and DLT with the perspective camera model fails. In summary, we present a method to obtain accurate camera as well as hand-eye calibration of the robot-camera system, localization of the tool, and reconstruction of the retina, all within the same pipeline. We use ex-vivo pig-eyes to validate our method. We provide detailed evaluation for each part, separately and in combination, showing several quantitative and qualitative results.

\section{Related Work}
\label{sec:related_work}
\textbf{CNNs for Landmark Localization:}
Convolutional Neural Networks (CNNs) have recently revolutionized many computer vision tasks. Image recognition on very large datasets such as ImageNet~\cite{Kri+12, SiZi15, He+16} is one of the most representative examples. Models initially trained on ImageNet can often be fine-tuned for a variety of tasks, thus producing state-of-the-art results, such as for object detection~\cite{Ren2015, Lin+17} and segmentation~\cite{Zha+16, Che+17}. Closest to our work are the CNN-based keypoint prediction methods, applied for Human Pose Estimation~\cite{NYD16, Pav+17, ToSz14}. Drawing inspiration from such methods, we use a CNN to directly regress the keypoints, and thus the 2D pose of the surgical tool. Pavlakos et al.~\cite{Pav+17a} use semantic keypoints to obtain the 6 degrees of freedom (DoF) pose of objects. Their pipeline is limited by the GPU memory, which enforces the authors to downsample the input images. In contrast, our method uses full-HD stereo images ($1080\times1920$), and we argue that keeping the input resolution is crucial for achieving accurate localization. Concurrent work~\cite{Laina2017} uses tool landmark detection for assisting segmentation. Different from that approach, we focus on instrument landmark detection to assist in 3D vision tasks, such as microscopic camera calibration, and robot-to-camera registration. The proposed method is also trained from scratch, meaning that we do not rely on pre-trained ImageNet weights that are difficult to acquire, and thus we are flexible in the network design. Our aim is to achieve real-time performance, which is usually not possible using very deep architectures~\cite{Laina2017}.

\textbf{Stereo Calibration and Reconstruction:}
There is an extensive literature on camera calibration for both stereo and monocular cameras~\cite{Hartley2003,Zhang2000}. Yet, the problem is different for microscope cameras. For the task of modeling the projection geometry, it is not clear which camera models and calibration methods provide the best results. For example, \cite{Ammi2009} considers a perspective camera model to calibrate a standard microscope while \cite{Chanwimaluang2006} considers an affine camera model for a fundus camera. Due to the special optical arrangement of the camera, the small size of the viewed object and its relatively large distance to the camera, we use the affine camera model. In~\cite{Chanwimaluang2006}, the authors propose to reconstruct the retinal surface using classical affine Structure from Motion (SfM)~\cite{Tomasi1992}, with a fundus camera. Such reconstruction is known only up to an unknown affine transform however, and the authors propose an upgrade to metric reconstruction by solving a highly non-linear cost function that requires a suitable initialization. The final retinal reconstruction is obtained only after fitting a spherical surface to the reconstructed points. 
In~\cite{Horaud1997}, the non-linear cost is avoided by using controlled robot motions so that the affine shape from factorization~\cite{Tomasi1992} can be used to formulate Linear Matrix Inequalities (LMI) for full affine calibration. Both works~\cite{Chanwimaluang2006,Horaud1997} rely on having an accurate affine reconstruction for calibration. In practice, affine factorization is challenging due to outliers, missing data bound to be present in long view sequences as required here, and hence may not give an accurate result.
We therefore propose to calibrate the affine cameras independently, using bundle adjustment accurately initialized by DLT before moving on to the reconstruction. This allows us to reconstruct surfaces online. We also do not require an {\em a priori} geometric model of the surface.

\section{Automatic Surgical Instrument landmark localization}
\label{sec:tooldetection}
\begin{figure*}[t]
\centering
\includegraphics[width=0.9\textwidth]{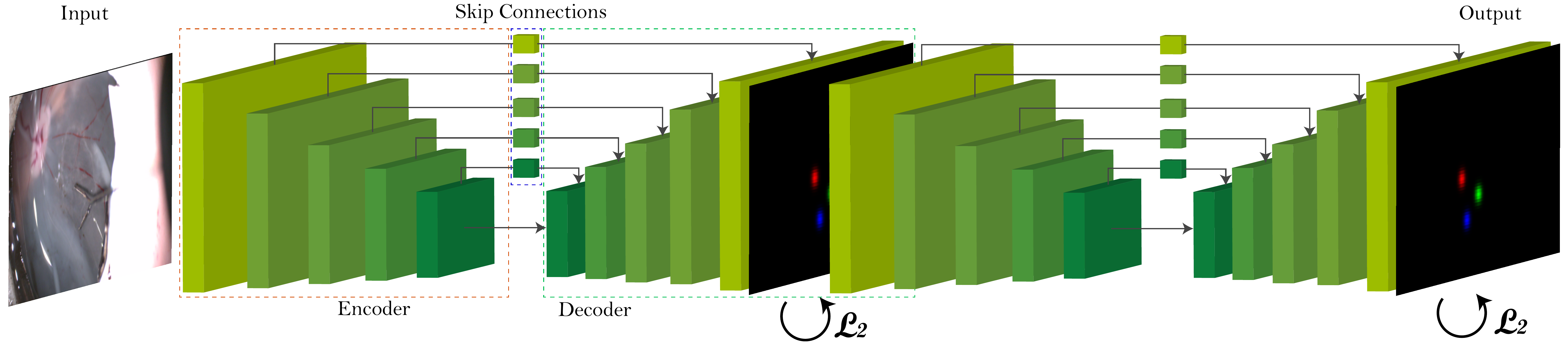}
\vspace{-2mm}
\caption{\textbf{Stacked Hourglass Network (SHN) architecture overview}. SHN is a fully convolutional architecture, which consists of multiple hourglass modules (two in the figure). Each hourglass is built by an encoder-decoder with skip connections. Each box represents a residual module~\cite{He+16}. Each hourglass is supervised by the ${l}_2$ loss.}
\label{fig:hourglass}
\vspace{-4mm}
\end{figure*}

\textbf{Number of keypoints:}
Before designing the localization algorithm, a careful consideration about the number of the landmark points required by subsequent parts of the pipeline is needed. In general, we require at least 3 non-collinear point correspondences to register two coordinate frames. We therefore design our CNN architecture to detect 3 keypoints in separate output channels. Note that the number of keypoints has a diminishing impact on the computational cost. In case of the surgical tool used for the retinal membrane peeling, we select the base of the tool, as well as the two tool tips as the landmark points of interest.

\textbf{CNN architecture:}
For keypoint localization, we re-implemented the Stacked Hourglass Network architecture (SHN)~\cite{NYD16}, which has been proven very effective for Human Pose Estimation. Human pose estimation is dominated by keypoint localization approaches, focusing on various joints and landmarks of the human body (eg. head, right shoulder, etc.). Inspired by this approach, we substitute the body landmarks by the instrument landmarks, which makes SHN suitable for our purpose, although the original task is substantially different. SHN is a fully convolutional architecture, that consists of convolutional, ReLU, and pooling layers. Its core component is an encoder-decoder network enriched with skip connections. SHN is created by stacking together multiple such components, in a way such that the output of a previous component is the input to the next. Like this, coarse and fine features are gradually interchanged by pooling and upsampling operations on the feature maps, which builds a powerful representation for dense prediction tasks. SHNs also make extensive use of residual blocks~\cite{He+16} and intermediate supervision~\cite{Lee+14} which further enhance their performance. Figure~\ref{fig:hourglass} provides an overview of the SHN architecture. For a more detailed description, we refer the reader to the original paper~\cite{NYD16}. We conducted a set of ablation experiments, with multiple architecture designs, where we concluded that the SHN architecture works best for the task of tool keypoint localization (Section~\ref{sec:abl-cnn}).

We formulate tool keypoint localization as a heatmap regression problem. Specifically, for each keypoint, we regress a heatmap swith its predicted location, as a separate channel of the CNN. We work with 3 keypoints and consequently 3 heatmaps. Our supervisory signal consists of the ground-truth locations, on top of which 2D Gaussians with standard deviation $\sigma$ are centered. Centering Gaussians around the keypoints improves stability during training, since they ensure a softer loss over slight mis-localized detections. We train to minimize the ${l}_2$ loss.
During inference, the peak activations in the final layer are considered the locations of the keypoints. Specifically, we obtain the location of the $k$th keypoint as:
\begin{equation}
\hat{\mathbf{u}}_{det,k} = \frac{1}{|\Delta|}\sum_{\Delta}\left(\hat{\mathbf{u}}_{max,k} + \Delta\right) p_{k}\left(\hat{\mathbf{u}}_{max,k} + \Delta\right)
\end{equation}
where $p_k(.)$ is the probabilistic activation of the $k$-th heatmap, $\hat x_{max}=\arg\!\max{\,p_k}$, and $\Delta$ is a small neighbourhood. In our case, we define $\Delta$ as a circular neighbourhood with radius $3\sigma$.
An example for the detected heatmaps is shown in Fig.~\ref{fig:qualitative}. The detected 2D keypoints, together with the corresponding 3D locations acquired by the robot kinematics are fed to the next stages of the pipeline: camera calibration, registration, and 3D reconstruction.

\section{Automatic Calibration and 3D Reconstruction}
\label{sec:calrecons}
\subsection{Stereo Camera Calibration Using Robot Kinematics}
The problem of stereo camera calibration refers to that of obtaining the intrinsics and pose (extrinsics) of the cameras. The stereo camera used in retinal microsurgery, such as the one in Fig.~\ref{fig:pipeline}, allow for a continuous adjustment of zoom and independent rotation of the cameras in a plane. Consequently, both extrinsics and intrinsics may change during the surgery. The standard way to calibrate a perspective camera is to use \cite{Zhang2000} on several images of a planar checkerboard pattern.
 
However, for the microscope cameras used in retinal microsurgery the projections are affine. This is because the distance from camera to object is orders of magnitudes larger than the object's size and the depth of field. Consequently, rays arrive almost parallel at the camera plane and perspective effects vanish. In such cases the equations of \cite{Zhang2000} are not well-conditioned and cannot be solved reliably.

\textbf{Automatic 3D-2D Correspondence Acquisition:}
In the case of robot-assisted surgery, we can exploit the fact that we are able to manipulate any surgical tool in 3D space while having instant position feedback computed using the robot kinematics. We therefore propose to tackle the problem of affine stereo calibration by relying on automatic detection of distinct keypoints on the robot tool in the image. Having access to synchronized real-time kinematics, we can automatically accumulate any desired number of 3D-2D correspondences $\mathcal{C}$. Note that, given the robot-assisted surgical procedure, we obtain the correspondences for free once we have the 2D tool landmark detections. 
\begin{equation}
\label{eq:correspondences}
	\mathcal{C} = \left\lbrace (\mathbf{x}, \mathbf{u}^{l},\mathbf{u}^{r})\in \,\mathbb{R}^3 \times \mathbb{R}^2 \times \mathbb{R}^2\right\rbrace_{t,k} \, \begin{array}{ll} t \in [1,n_t]\\k \in [1,n_k] \end{array}
\end{equation}
While observing a sequence of $n_t$ frames with a static camera pair, we detect $n_k$ keypoints of the moving tool in each frame $t$, resulting in a set of $|\mathcal{C}|=n_k n_t$ correspondences. For each correspondence in $(\mathbf{x} ,\mathbf{u}^l,\mathbf{u}^r )_i \in \mathcal{C}$, $\mathbf{x}$ is the tool 3D landmark expressed in the robot coordinate system while we refer to the corresponding 2D keypoints on the images as $\mathbf{u}^l$ and $\mathbf{u}^r$ for the left and the right camera. We use the subscript $i$ as $\mathbf{x}_i$, $\mathbf{u}^{l}_{i}$ or $\mathbf{u}^{r}_{i}$ to denote the $i$-th 3D point in $\mathbf{C}$ and its projection on the left and right image, respectively.

\textbf{Joint Affine Stereo Pair Calibration:}
We now formulate camera calibration as a problem of fitting an affine camera to model the image projections from given 3D points in the robot reference frame. Since we can control the robot, we make sure that a sufficient 3D volume is covered with point correspondences, to maximize the calibration accuracy. We are interested in an online stereo system that triangulates and reconstructs surfaces close to real-time from a pair of stereo images. Thus we deviate from the standard calibration methods based on affine reconstruction~\cite{Chanwimaluang2006,Horaud1997} and triangulate Euclidean shapes directly using calibrated cameras.
The affine camera projection is modeled by the projection matrix $\mathsf{M}\in \mathbb{P}_{\text{Affine}}\subset\mathbb{R}^{2\times4}$ as $\mathbf{u}=\mathsf{M} \!\left[\mathbf{x}^\top \, 1\right]^\top$.

In order to jointly calibrate the stereo pair $\mathsf{M}^l,\,\mathsf{M}^r$, while accommodating for noise in the 2D detections and 3D measurements, we write the following energy to robustly minimize reprojection errors in a bundle adjustment fashion:
\begin{align}
\label{eq:bundleAdjustment}
  \min\limits^{}_{\mathsf{M}^l, \mathsf{M}^r, \tilde{\mathbf{x}},\tilde{\mathbf{u}}^l,\tilde{\mathbf{u}}^r} & \mathcal{E}_{\Pi}(\mathsf{M}^c, \tilde{\mathbf{x}},\tilde{\mathbf{u}}^c) + \sigma_{u}^{-1} \mathcal{E}_{\Theta}(\tilde{\mathbf{u}}^c) + \sigma_{x}^{-1} \mathcal{E}_{\Phi}(\tilde{\mathbf{x}}) \nonumber\\
  \text{subject to}, \quad &\mathsf{M}^c \in \mathbb{P}_{\text{Affine}}, \quad c \in \{l,r\}\nonumber\\
  &\mathcal{E}_{\Pi}(\mathsf{M}^c, \tilde{\mathbf{x}},\tilde{\mathbf{u}}^c)=\frac{1}{2} \sum_{c} \sum_i (\tilde{\mathbf{u}}^{c}_{i}- \mathsf{M}^c \tilde{\mathbf{x}}_i)^2 \nonumber\\
  &\mathcal{E}_{\Theta}(\tilde{\mathbf{u}}^c)=\frac{1}{2} \sum_{c} \sum_i (\tilde{\mathbf{u}}^{c}_{i}-\mathbf{u}^c_i)^2 \nonumber\\
  &\mathcal{E}_{\Phi}(\tilde{\mathbf{x}})=\frac{1}{2} \sum_i (\tilde{\mathbf{x}}_{i}-\mathbf{x}_{i})^2.
\end{align}  
The minimization problem in Eq.~\eqref{eq:bundleAdjustment} is essentially the bundle adjustment for an affine camera. The first term describes the reprojection error. The last two terms model the uncertainty in the measurements as Gaussians with standard deviations $\sigma_{u}$, $\sigma_{x}$. We assume different intrinsics for each of the stereo camera pair and jointly optimize for the camera parameters $\mathsf{M}^{l,r}$, the 3D point positions $\tilde{\mathbf{x}}_i$, as well as for the 2D projections $\tilde{\mathbf{u}}^{l,r}_{i}$. Eq.~\eqref{eq:bundleAdjustment} is optimized using a gradient-based interior-point technique.

\textbf{Robust DLT for affine camera projection:}
In order to initialize the non-linear problem in Eq.~\eqref{eq:bundleAdjustment} with a feasible configuration, we perform an affine Direct Linear Transform (DLT) on each camera separately. Writing down the affine projection for each point gives us the following system of equations for each camera:
\begin{equation}
\label{eq:multiviewproj}
	\begin{bmatrix} \mathbf{u}_1 & \hdots & \mathbf{u}_n\end{bmatrix} = \mathsf{M} \begin{bmatrix} \mathbf{x}_1 & \hdots & \mathbf{x}_n \\ 1 & \hdots & 1\end{bmatrix}.
\end{equation}
We solve for $\mathsf{M}$ in Eq. \eqref{eq:multiviewproj} by using the DLT algorithm~\cite{Hartley2003} modified for affine projections. The problem becomes that of a linear least squares (LLS) that requires a minimum of $n=4$ non-coplanar points. To tackle outliers and noise in 2D detections, we use Random Sample and Concensus (RANSAC) to estimate the projection matrix with Eq.~\eqref{eq:multiviewproj}. Although camera distortion could easily be included, this is not used for the sake of stability and better robustness against noise. Including distortion parameters also increases the number of minimum points needed by RANSAC, as well as the number of parameters in Eq.~\eqref{eq:bundleAdjustment}, outweighing the advantages of a more complex model.

\textbf{Affine camera resection:}
For the perspective camera, resection refers to the decomposition of the projection matrix into the intrinsic calibration matrix and the 6 DoF pose of the camera. In the affine model, the problem is similar but only two rows of rotation and two translation components exist in the affine projection matrix. Consequently, the decomposition of the affine projection matrix is expressed as: \vspace{-2mm}
\begin{equation}
\label{eq:camresection}
	\mathsf{M} = \mathsf{K} \begin{bmatrix}
		\mathbf{r}_1^\top & t_1 \\
		\mathbf{r}_2^\top & t_2
	\end{bmatrix}, \quad 	\mathsf{K} = \begin{bmatrix}
		\alpha_x & s \\ 0  & \alpha_y
	\end{bmatrix}
\end{equation}
where $\mathbf{r}_1^\top$ and $\mathbf{r}_2^\top$ are the first and second rows of a rotation matrix; $t_1$ and $t_2$ are the translation components, and $\mathsf{K} \in GL_2(\mathbb{R})$ is the intrinsic affine calibration matrix. $\mathsf{K}$ is found by QR factorization of the projection matrix $\mathsf{M}$. This gives the intrinsic calibration as well as the pose of each camera with respect to the robot reference frame except for the translation along the optical axis. The optimal camera parameters follow from Eq.~\eqref{eq:bundleAdjustment} while enforcing $s=0$ for stability. The intrinsics and pose parameters estimated from the bundle adjustment are used to recompute the camera projection matrices.

\subsection{Stereo Matching and Reconstruction}
The standard pipeline for stereo reconstruction with calibrated perspective cameras consists of dense disparity computation and depth map estimation by triangulation. The lighting used in retinal microsurgery often contaminates the images with specularities as well as other reflections, however. We therefore opt for a semi-dense matching method such as Deep Matching \cite{Revaud2013}. We filter out outliers based on the epipolar geometry derived from the affine fundamental matrix obtained during calibration.
We then triangulate the matched points using the two affine projection matrices $\hat{\mathsf{M}}^{l}$ and $\hat{\mathsf{M}}^{r}$ for the stereo pair. Triangulation is possible because our estimated projection matrices are accurate. This directly gives us the 3D points of the observed surface in the Euclidean robot reference frame.

\textbf{Robust surface estimation:}
The 3D points obtained from the triangulation contain outliers and noise due to two reasons. First, outlier removal using the epipolar geometry cannot reject all outliers in the stereo matches. Second, the affine triangulation is sensitive to noise naturally present in the 2D correspondences. In such case using a surface prior model such as a sphere for the retina~\cite{Chanwimaluang2006}, can make the reconstruction better, but such a surface constraint may be too limiting. In the case of the open-sky pig eyes used for our tests, the retinas are far from spherical and can be of any smooth shape. We therefore propose to fit a single surface using Bicubic B-Splines (BBS) \cite{Wahba1990}.
We use the image as the parametrization space for representing the surface. The surface $\Psi: \mathbf{u} \to \hat{\mathbf{x}}$ is thus a function of the image points $\mathbf{u}\in \Omega$ and the spline coefficients $\mathbf{c}\in\mathbb{R}^{2n_c}$, where $n_c$ is the number of spline coefficients used to represent the surface. Consider there are $n_r$ 3D points with the same number of 2D image correspondences. We express the surface reconstruction problem as:
\begin{equation}
\label{eq:splinesl1}
	\hat{\mathbf{c}} = \argmin_{\mathbf{c}} \sum_{i=1}^{n_r} \left\lVert \Psi(\mathbf{u}_i;\mathbf{c}) - \mathbf{x}_i \right\lVert_1 + \mu \int_\Omega \norm{\frac{\partial^2} {\partial \mathbf{u}^2} \Psi(\mathbf{u};\mathbf{c})}^2_2
\end{equation}
Eq.~\eqref{eq:splinesl1} consists of a data term under the ${l}_1$-norm as well as a regularizer which penalizes very high frequency changes over the surface. The two terms are balanced by a hyperparameter $\mu \in \mathbb{R}^+$. We choose the ${l}_1$-norm to obtain a more robust surface fitting~\cite{Chhatkuli2017}. We also reject points where the ${l}_1$-norm of the data term exceeds a certain threshold $\epsilon$. We then re-estimate the surface by solving Eq.~\eqref{eq:splinesl1} with the remaining points. The single iteration of point rejection and surface re-estimation gives a surface that is smooth and largely free of the reconstruction noise.

\subsection{Registration}
\label{sec:registration}
We define registration as the transformation of the camera pose and reconstructions to the robot coordinate frame.
This is necessary because the stereo microscope (and the mounted cameras) may be moved during
its use. Such motion can be measured from the images by the Perspective n-Point (PnP) method~\cite{Lepetit2009}. However, PnP cannot be used with the affine camera and we therefore compute registration using the reconstruction of the tool landmark positions and their positions measured by the robot kinematics.

Consider $(\mathsf{R}^c, \mathbf{t}^c)$, $c\in\{l,r\}$ to be the 6 DoF pose of the camera $c$ with respect to its initial position, where  $\mathsf{R}^c\in SO_3$ is the rotation and $\mathbf{t}^c\in \mathbb{R}^3$ is the translation undergone by the microscope cameras.
We then express the registration problem as: \vspace{-4mm}
\begin{equation}
\label{eq:registration}
\min_{\mathsf{R}^c,\ \mathbf{t}^c}\sum_{i=1}^{n_k n_f}\left\lVert \mathsf{R}^{c}\hat{\mathbf{x}}_i +\mathbf{t} - \mathbf{x}_i^{\text{gt}}\right\rVert_2
\end{equation}
where $\hat{\mathbf{x}}_i$ is the $i$th triangulated tool landmark 3D point and $\mathbf{x}_i^{\text{gt}}$ is the 3D $i$th tool landmark point as measured by the robot's measurement system. Eq.~\eqref{eq:registration} is a well-studied problem and can be solved linearly using only three non-collinear points. In practice, a more accurate pose estimate is obtained by using multiple frames and accumulating 3D-3D correspondences, assuming a static camera within this time window.

\section{Experiments}
\label{sec:experiments}
\subsection{Dataset}
In order to train the deep network for keypoint localization, we manually annotated sequences of stereo images for the tool of interest with 3 keypoint locations. Such annotations are acquired with minimal effort, since for each stereo pair only 6 mouse-clicks are necessary. Apart from the manual 2D annotations, we acquired the 3D locations of the keypoints, from the kinematics of the robot. The dataset consists of 10 sequences, acquired from different pig eyes, with both artificial movements that help calibration and realistic movements performed by a surgeon. It includes more than 1500 full HD images and their labels. We limit acquisition to one type of tool, since our aim is accuracy rather than generalizing to different ones. The method itself is easily adaptable to other types of surgery, and tools with different landmarks. The dataset will be publicly released to ease further research.

\subsection{Evaluation of Keypoint Localization}
\label{sec:abl-cnn}
\textbf{Training details:}
For keypoint localization, we split the data into training and testing sets, and train the SHN model for 150 epochs. We use 7 sequences for training, and 3 for testing. Results for localization are reported for all images of the testing set. We use RMSProp~\cite{NYD16} with $\alpha=0.99$ and zero momentum. The initial learning rate is set to $5\cdot 10^{-5}$, and is adapted by RMSProp for each of the layers. We use a standard deviation of $\sigma=5$ for the 2D gaussians centered on the keypoints. To avoid overfitting, we use extensive data augmentation that consists of random rotations $\left[ -30^\circ, 30^\circ\right]$, and zooming $\left[ 0.75\times, 1.25\times \right]$. The images of the training set are randomly permuted, and a single model is trained for both the left and the right camera. For all our experiments, we train the models from scratch, in less than 4 hours with an NVidia Titan-X GPU. During testing, our batch contains both left and right images of the stereo camera. We note that during inference the CNN processes the images in higher than real time speed. Real time performance is especially important for 3D registration when the camera moves, and thus we keep all our experiments above the threshold of $30Hz$. We found that in practice, a stack of 2 Hourglasses is a fair compromise of speed and accuracy, when processing full-HD ($1920\times 1080$) images. Common models pre-trained on ImageNet~\cite{SiZi15, He+16, Laina2017} are much more memory and computation intensive, not allowing to process images at such resolution, let alone in real-time.

\textbf{Evaluation metric:}
For evaluation of the 2D keypoint localization, we use the Percentage of Correct Keypoints (PCK) measure (also referred to as KBB~\cite{Laina2017}). In PCK, a detection is considered correct, if it falls `near enough' to the label. The threshold is computed as a percentage of the distance of the tool-tip from its base.

\textbf{Network architecture ablation:}
In order to decide on the final CNN architecture, we conduct an ablation experiment to show the importance of each of the used components. Starting from an encoder architecture like the ones used for image classification (without the fully connected layers), we observe poor performance (8.4\% PCK) due to the heavily downsampled output. Adding the decoder architecture immediately solves this problem (75.1\%). Skip connections and a second hourglass boost the overall performance further (95.2\% and 99.6\%). Substituting the convolutional modules by residual ones gives diminishing returns.

\begin{table}[h]
\centering
\small
\caption{\emph{CNN architecture ablation.} Various CNN architectures tested for keypoint localization and their quantitative contributions to the result.}
\label{tab:cnn_architecture}
\resizebox{0.9\linewidth}{!}{%
\begin{tabular}{l|ccccc}
\toprule
Architecture  & Encoder  & +Decoder   & +Skip Connections    & +Stacked     & +Residual     \\
\midrule
PCK@0.05 & \ 8.4 &    \ 75.1 &    \ 95.2 &    \ \bf 99.6 &    \ \bf 99.7     \\
\bottomrule
\end{tabular}}
\vspace{-0.5em}
\end{table}

\begin{figure}[h]
\centering
\resizebox{.85\linewidth}{!}{\begin{tikzpicture}[/pgfplots/width=1\linewidth, /pgfplots/height=0.64\linewidth, /pgfplots/legend pos=south east]
    \begin{axis}[ymin=0,ymax=100,xmin=0,xmax=.2,enlargelimits=false,
        xlabel=Threshold,
        ylabel=PCK \%,
		font=\scriptsize,
        grid=both,
		grid style=dotted,
        xlabel shift={-2pt},
        ylabel shift={-5pt},
        legend columns=1,
        legend style={/tikz/every even column/.append style={column sep=3mm}},
        minor ytick={0,25,...,110},
        ytick={0,10,...,110},
		yticklabels={0,10,20,30,40,50,60,70,80,90,100},
	    xticklabels={0.0, 0, .05, .1, .15, .2},
        legend pos= north east
        ]
        
        \addplot+[smooth,green,mark=none, line width=1] table[x=range,y=values] {data/hg2_highres_Calibration_1.txt};
        \addlegendentry{$1080p$}
        \label{fig:pck:highres}
        
        \addplot+[smooth,black,mark=none, line width=1] table[x=range,y=values] {data/hg2_Calibration_1.txt};
        \addlegendentry{$480p$}
        \label{fig:pck:midres}
        
        \addplot+[smooth,red,mark=none, line width=1] table[x=range,y=values] {data/hg2_lowres_Calibration_1.txt};
        \addlegendentry{$(480\times640)$}
        \label{fig:pck:lowres}
        
        \addplot+[smooth,blue,mark=none, line width=1] table[x=range,y=values] {data/hg2_verylowres_Calibration_1.txt};
        \addlegendentry{$(256\times448)$}
        \label{fig:pck:verylowres}

    \end{axis}
\end{tikzpicture}}
\vspace{-3mm}
   \caption{\textbf{Tool localization accuracy}: The PCK accuracy measure as a function of the maximum tolerance, for different input image resolutions. Tolerance is normalized by the size of the instrument tip.}
   \label{fig:qual_vs_thres}
   \vspace{-3mm}
\end{figure}
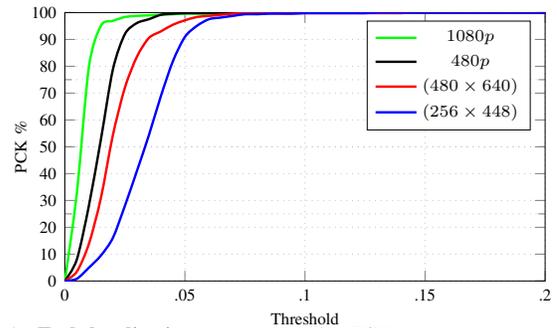

\begin{figure}
\centering
\includegraphics[width=.88\linewidth]{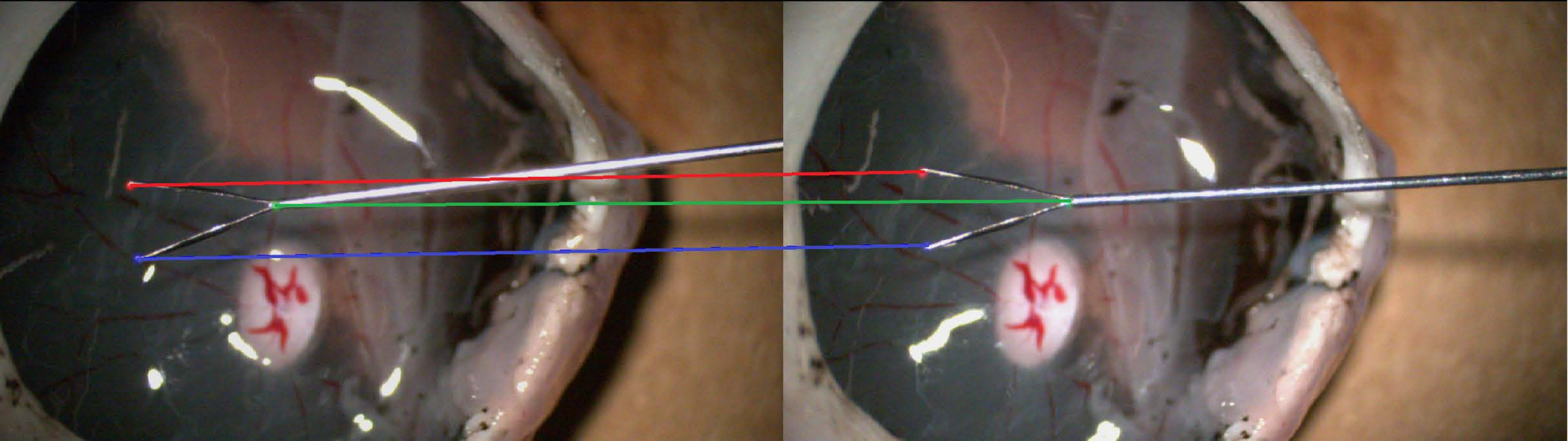}\\
\resizebox{.90\linewidth}{!}{%
	  \setlength{\fboxsep}{0pt}
      \fbox{\includegraphics[width=0.5\linewidth]{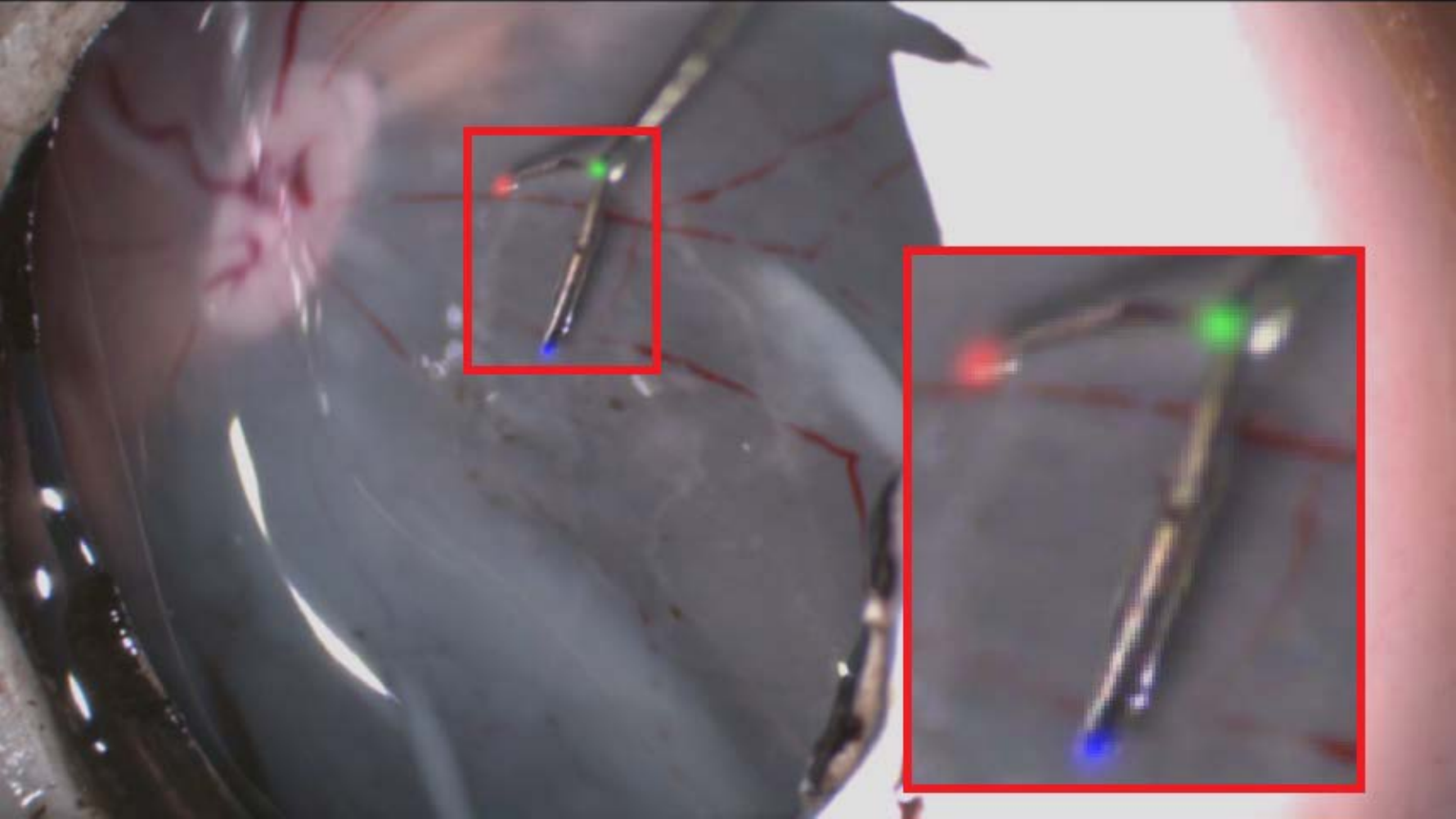}}
      \fbox{\includegraphics[width=0.5\linewidth]{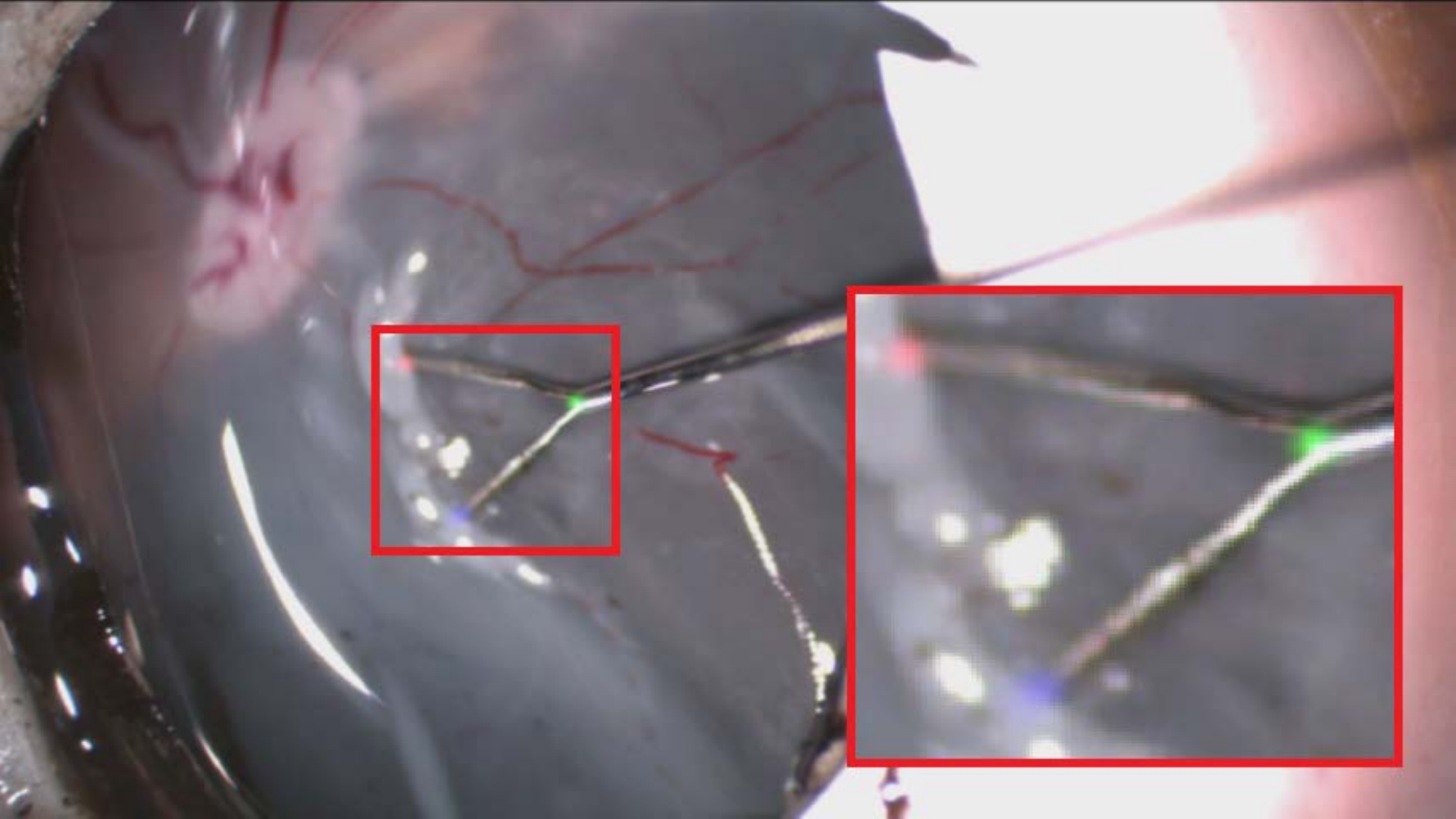}}
      }\\[-1mm]
\caption{\textbf{Qualitative results}: Keypoint localization on an example stereo image pair (top) and more qualitative example in another scenario (bottom).}
\label{fig:qualitative}
\vspace{-3mm}
\end{figure}

\textbf{Input image resolution:}
Fig.~\ref{fig:qual_vs_thres} illustrates the PCK measure as a function of the threshold for the accepted mis-localization, for various input image resolutions. For full-HD images, above the threshold of ~1\%, almost all detections are correct. The same accuracy is obtained for images of $480 \times 640$, for a threshold 6 times larger. Small errors in 2D lead to larger errors in 3D, so we argue that accurate 2D localization is crucial for the next steps of the pipeline, such as calibration (Fig.~\ref{fig:calib_resolutions}). Fig.~\ref{fig:qualitative} shows some qualitative examples of keypoint localization, obtained for high resolution images.

\textbf{Timing:}
Table~\ref{tab:timing} shows the execution rate of the CNN when the input resolution is varied. The timing regards the forward pass and the post-processing to obtain the locations of the landmarks from the heatmaps, for a batch of 2 images (left and right). Although we sacrifice execution speed for accuracy by using full HD images, the landmark localization remains faster than real-time (30Hz) at all resolutions. All experiments were conducted on a NVidia Titan-X GPU.

\begin{table}[h]
\small
\centering
\caption{\emph{Execution Times}: Performance as a function of the input image resolution. All models achieve better than real-time performance.}
\label{tab:timing}
\resizebox{0.9\linewidth}{!}{%
\begin{tabular}{l|ccccc}
\toprule
Resolution  & $1080p$  & $480p$   & $480 \times 640$    & $256 \times 448$   \\
\midrule
Frequency (Hz) & \ 35 &    \ 77 &    \ 95 &    \ \bf 140 &   \\
\bottomrule
\end{tabular}}
\vspace{-1em}
\end{table}

\subsection{Evaluation of Calibration}
\begin{figure}
\newcommand{\myHeight}{2cm}
\centering
\begin{minipage}[t][\myHeight][c]{1\columnwidth}
\begin{minipage}[t][\myHeight][c]{0.5\textwidth}%
  \centering
  \includegraphics[width=1\textwidth]{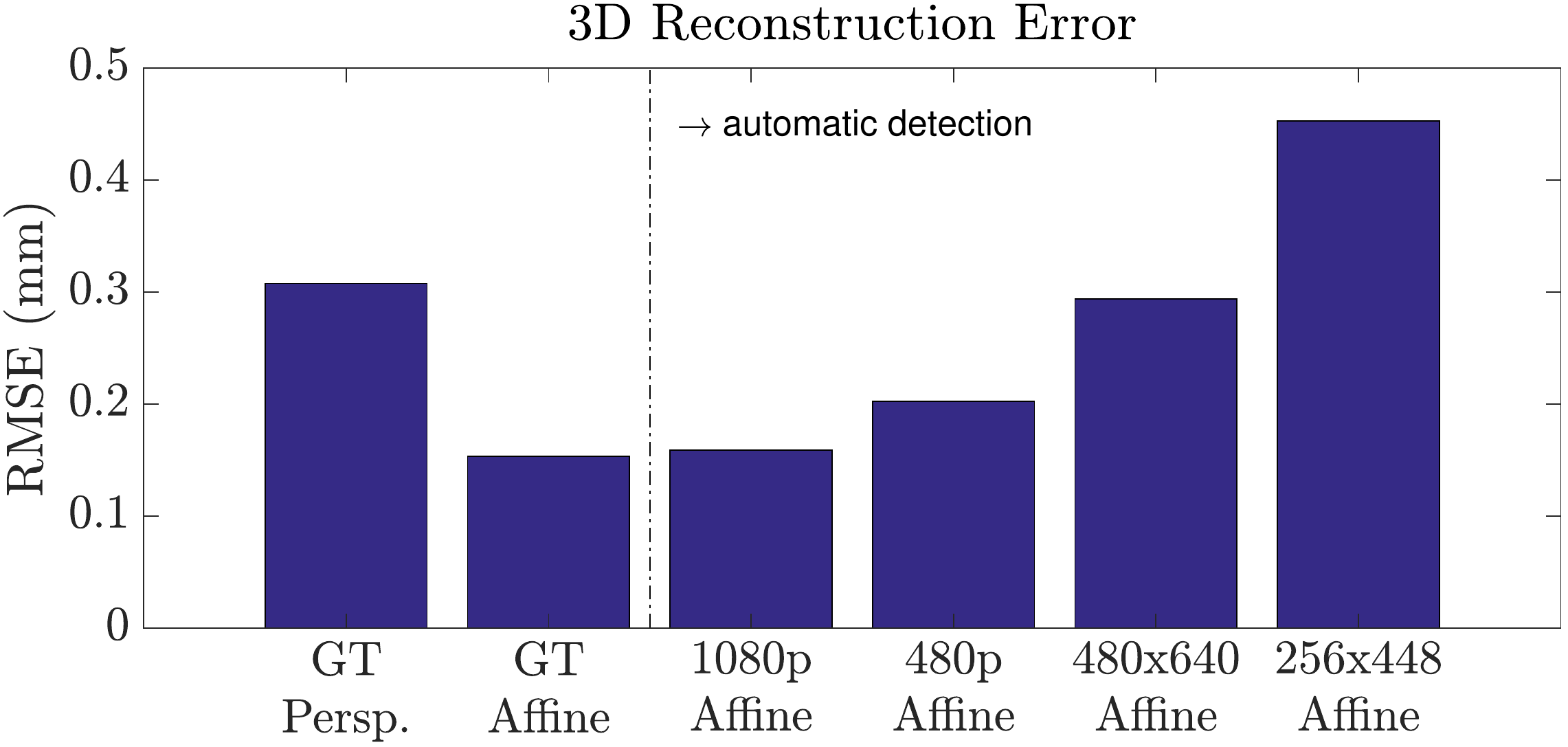}
\end{minipage}%
\begin{minipage}[t][\myHeight][c]{0.5\textwidth}%
  \centering
  \includegraphics[width=1\textwidth]{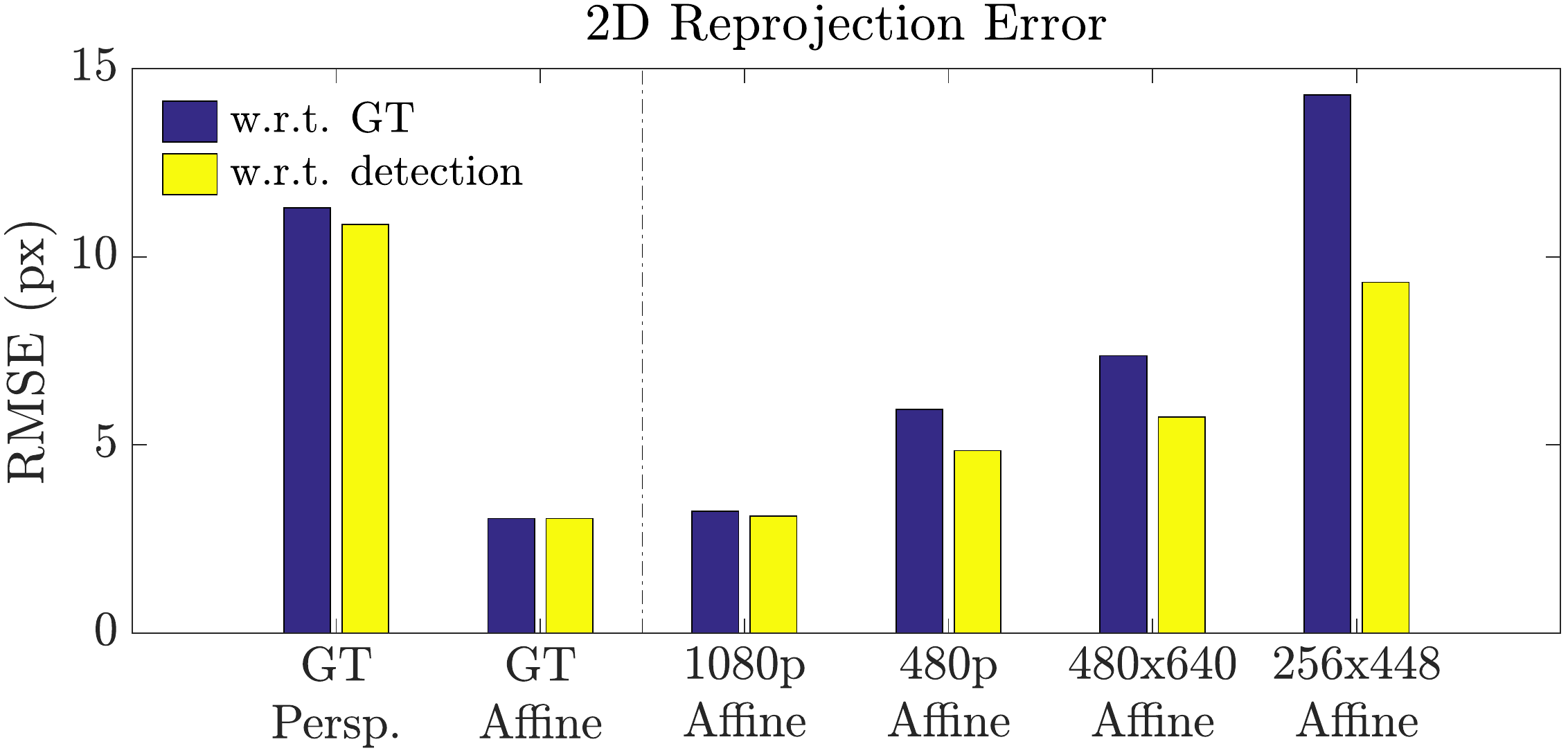}
\end{minipage}%
\end{minipage}%
\caption{\textbf{Calibration accuracy.} We calibrate the stereo cameras using annotated tool keypoints (GT), and tool keypoint detections at four different image resolutions. The figures illustrate the 3D triangulation error (left) and 2D reprojection errors (right) for perspective and affine calibration. Clearly, the affine model performs better.}
\label{fig:calib_resolutions}
\vspace{-1em}
\end{figure}

As to the calibration, we first investigate the influence of the image resolution used for keypoint detection. Fig.~\ref{fig:calib_resolutions} shows the 3D reconstruction error and the 2D reprojection error for calibrations based on ground truth (GT) annotations and based on detections from four different image resolutions. Note that for 1080p we obtain virtually the same calibration quality as by using the manual annotation. As expected, the triangulation error as well as the reprojection error increase with lower resolutions. To validate the fitness of the affine camera model, we compare results with a perspective camera model calibrated with DLT from the full resolution annotations. This yields a much more unstable result compared to the affine model on the same data. Additionally, decomposition of the perspective projection matrix is not possible due to the influence of the large focal length on the matrix conditioning.

Using our automatic affine calibration, we performed a second experiment to gauge reconstruction accuracy for a known planar calibration object. Instead of relying on the robot kinematics, we analyzed calibration accuracy by reconstructing points on a checkerboard with 0.5 \si{\milli \meter} squares. To minimize mismatching and correspondence noise we use manual correspondences refined by a corner detector. In this optimal setup, we observe a Root Mean Square Error (RMSE) of 25.479 \si{\micro \meter} in the reconstruction.

\begin{figure*}
\centering
\includegraphics[width=0.9\textwidth]{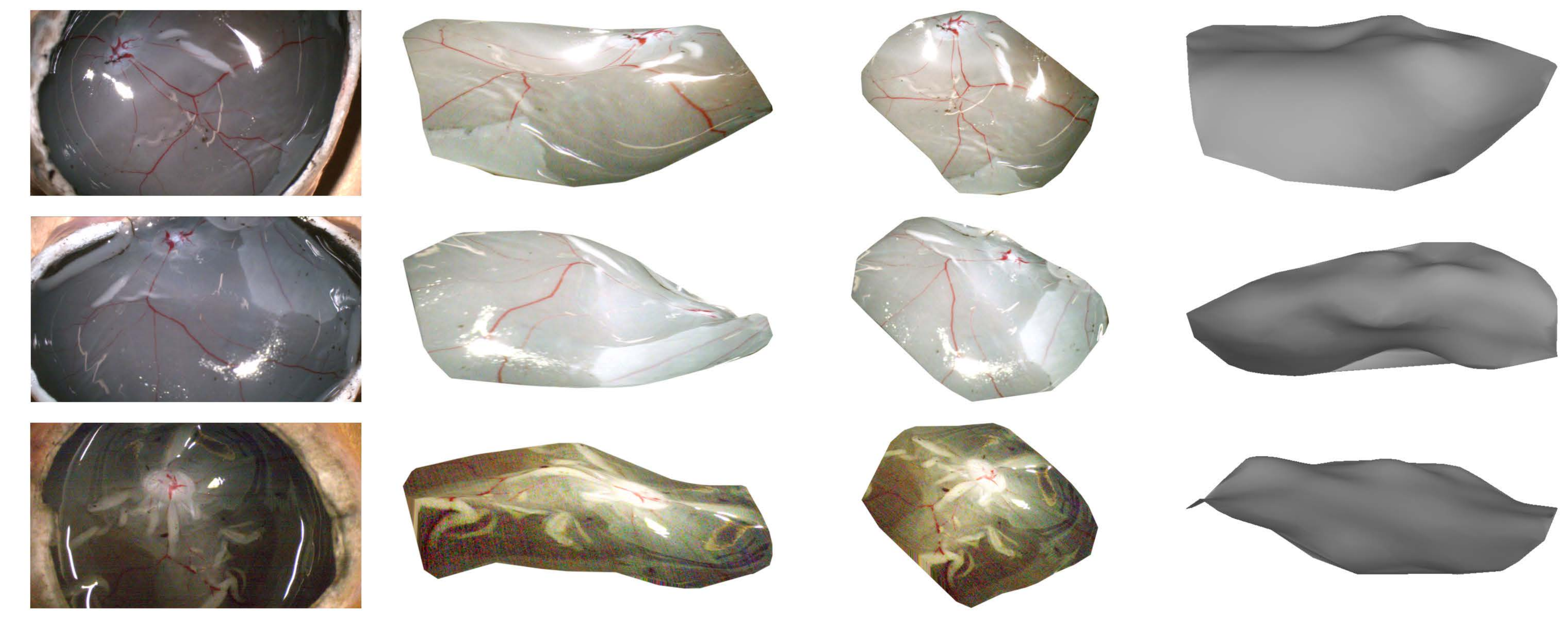}
\caption{\textbf{Pig Eye Reconstruction.} Each row shows the result of our reconstruction method for a different pig eye.}
\label{fig:recon_pigeyes}
\end{figure*}

\begin{figure*}
\centering
\includegraphics[width=0.9\textwidth]{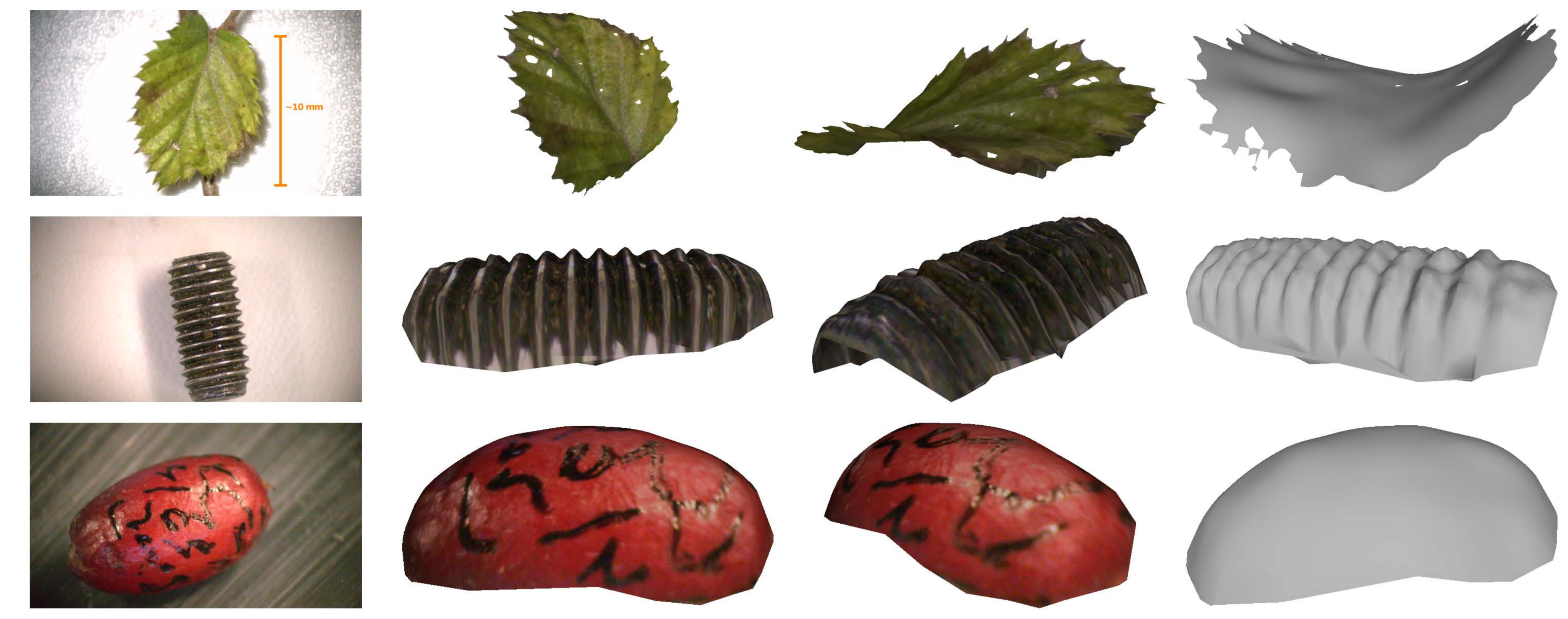}
\caption{\textbf{Generic Object Reconstruction.} Each row shows the result of our reconstruction method for one object.}
\label{fig:recon_objects}
\end{figure*}

\subsection{Retinal Reconstruction and Tool Registration}
\begin{figure}
\centering
\includegraphics[width=0.7\columnwidth]{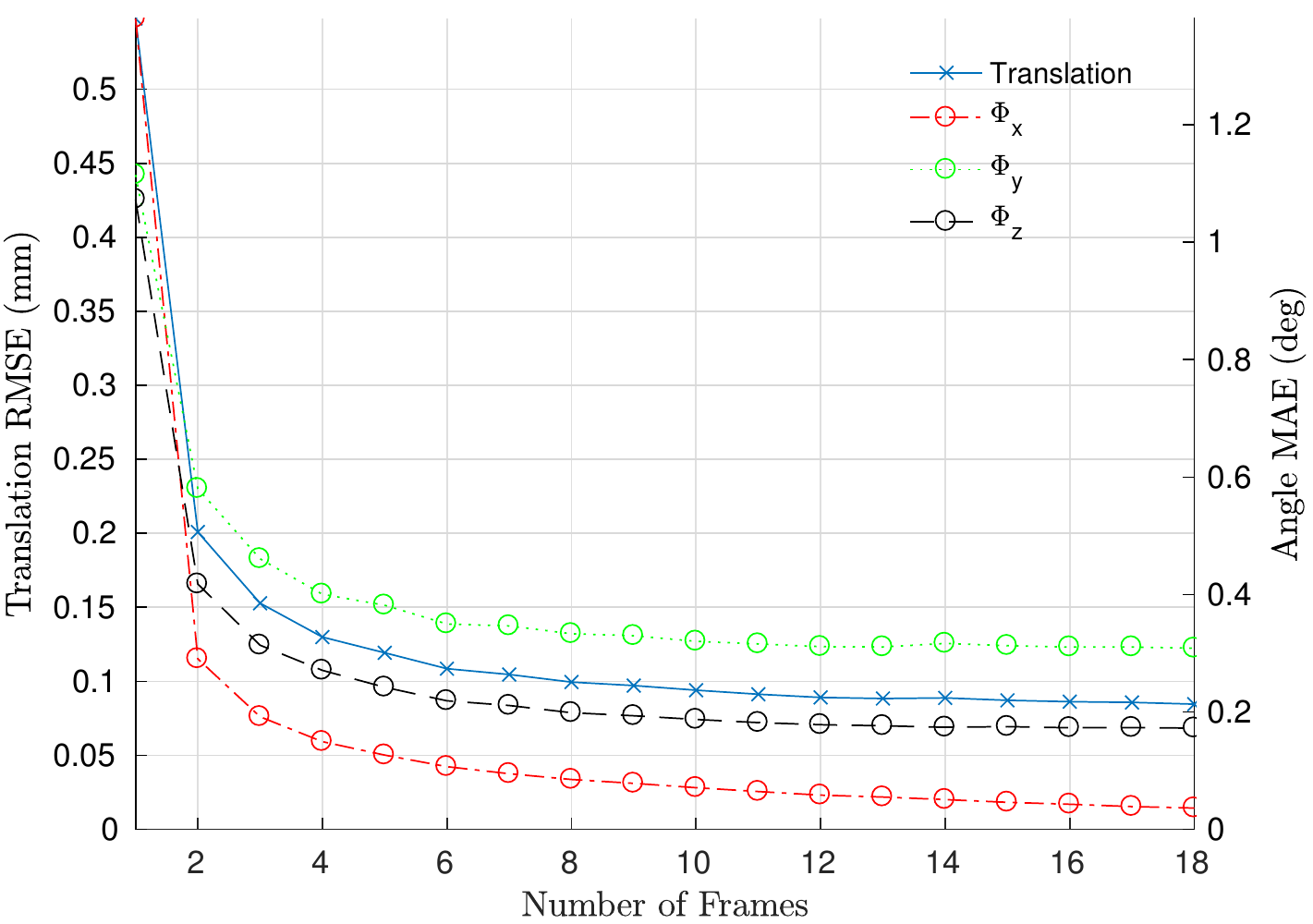}
\caption{\textbf{Tool Registration.} Translational and rotational errors of the estimated alignment of the tool w.r.t. the camera frame. We express rotational errors with Euler angles ($\Phi_{X}$, $\Phi_{Y}$, and $\Phi_{Z}$).}
\label{fig:registration_error}
\vspace{-2em}
\end{figure}

We reconstruct three open-sky pig eye sequences, each one for a different eye. The left camera image and the corresponding reconstructions are shown in Fig.~\ref{fig:recon_pigeyes}. To qualitatively evaluate the reconstructions, we show some of everyday objects in Fig.~\ref{fig:recon_objects}. Like the pig-eye the objects are roughly 1cm in size. The reconstructions of the screw and leaf are particularly interesting because this shows that we can get high and low frequency surface aspects.
Finally, we evaluate the registration using reconstructed tool points. Since there is no ground-truth label regarding the relative positions of the cameras, we synthetically move them by changing their projection matrices and measure the new pose using Eq.~\eqref{eq:registration}. We use one to several frames of the moving tool to measure the pose accuracy. Using $n_k=3$ keypoints, we achieve an error below 150 \si{\micro \meter} after about 3 frames as shown in Fig.~\ref{fig:registration_error}. This shows that we are able to quickly recover from 3D tracking failure in case the camera undergoes a change in pose by monitoring the consistency of the transformation over time. Note that the kinematics of our robot achieve accuracy of approximately 10 \si{\micro \meter}, whereas the diameter of a targeted vessel can range between 50 and 300 \si{\micro \meter}~\cite{Est+15}. The online reconstruction pipeline in our proof-of-concept implementation runs at about 5Hz speed, the main bottleneck being the DeepMatching~\cite{Revaud2013} method. Note that, although the speed may be increased further in the running system, currently envisioned applications do not necessarily require real-time reconstruction speed.

\section{Conclusions and Discussion}
\label{sec:conclusions}
This paper presented a new method for camera calibration, 3D reconstruction and registration, from automatically detected keypoints on a robotic end effector. Specifically, we use the 3D locations of the tool and the corresponding 2D locations on the acquired stereo images to establish correspondences, and initialize an affine bundle adjustment with the DLT method. We proposed a Stacked Hourglass CNN to detect the keypoints, which results in a very accurate and fast localization of the landmarks. We applied our method to robot-assisted eye surgery, where 3D processing is complicated by various issues with microscope camera imaging and the quality of the acquired data. We validated each component of our pipeline independently and in combination. We created and released a database that can facilitate training CNNs for the task, and we show quantitative and qualitative results for all the steps of our algorithm. Results show high quality keypoint localization, 3D reconstruction, and registration, all in the context of a single pipeline.

Although our pipeline works well in open-sky eyes, and is easily adjustable to different types of robotic surgery, in-vivo retinal surgery is a special case which comes with many challenges. For example, one needs to take into account further distortions from the lens of the examined eye. Additionally, we may encounter strong illumination changes, bleeding, the fact that the surgeon's movements are limited by the incision point of the tool, and last but not least, the fact that such data are difficult to acquire. Nevertheless, our method is the first to tackle 3D reconstruction and registration in affine microscopic cameras, and motivates future work in this direction.

\section*{ACKNOWLEDGMENTS}

Research funded by the EU Framework Programme for Research and Innovation Horizon 2020 (Grant No. 645331, EurEyeCase). We thank NVidia Corporation for donating the GPUs used in this project, Georgios Pavlakos and Danda Pani Paudel for their insightful comments, as well as Carlos Eduardo Porto de Oliveira for the visualizations.


{\small
\bibliographystyle{ieee}
\bibliography{biblio}
}

\end{document}